\documentclass[lettersize,journal]{IEEEtran}

\usepackage{amsmath,amssymb}
\usepackage{graphicx}
\usepackage{booktabs}
\usepackage{multirow}
\usepackage{siunitx}
\usepackage{url}
\usepackage{xcolor}
\usepackage{hyperref}
\hypersetup{
  colorlinks=true,
  linkcolor=black,
  citecolor=black,
  urlcolor=black,
}

\begin{document}

\title{Literary Emotions in Motion: A Soft Robotics Installation for Tactile Storytelling\\}
 
\author{Carolina Silva-Plata$^{*}{\dagger}$, Abraham Villavicencio-Carmona$^{\S}$, Miguel Silva Plata$^{\diamond}$, Stefan Escaida$^{\ddagger}$, Ruben Fernandez$^{\dagger}$%
\thanks{$^{*}$ Corresponding author.}%
\thanks{$^{\dagger}$ Department of Mechanical Engineering, University of Chile, Santiago, Chile.}%
\thanks{$^{\S}$ Independent Researcher, Santiago, Chile.}%
\thanks{$^{\diamond}$ Bolivian Catholic University, La Paz, Bolivia.}%
\thanks{$^{\ddagger}$ Institute of Engineering Sciences, University of O'Higgins, Rancagua, Chile.}%
}

\maketitle
\begin{abstract}
Soft robotics is increasingly explored in artistic contexts, where tactile interaction provides audiences with embodied engagement beyond visual or auditory signals. This work presents an interactive installation that maps semantic emotion analysis of narrative text into variable stiffness of soft pneumatic modules. A natural language model identifies two dominant emotions from a predefined set of six, driving the inflation of seven hexagonally arranged soft actuators. The central actuator represents the primary emotion, while the surrounding ones express the secondary. We develop and mechanically characterize silicone actuators, called soft modules, featuring a thin membrane layer, demonstrating how this morphological control expands the achievable stiffness range while preserving simplicity and low-cost fabrication. A user study with ten participants further evaluates how multisensory coupling of stiffness and LEDs intensity influences emotional perception. The results suggest that stiffness modulation accompanied by color change can support emotionally meaningful and engaging tactile interaction in soft robotic installations.

\end{abstract}

\begin{IEEEkeywords}
Soft robotics, variable stiffness, affective computing, expressive robotics.
\end{IEEEkeywords}

\section{Introduction}
Soft robotics has emerged as a promising paradigm for creating safe, adaptive, and expressive installations capable of interacting with humans. Its intrinsic compliance and deformability make it particularly suited for delicate manipulation, embodied expression, and human–robot interaction scenarios where safety and comfort are critical. Recent work at the intersection of soft robotics, affective computing, and the arts has shown that mechanical softness, organic morphologies, and dynamic shape transformations can trigger emotional responses and enrich interactive experiences \cite{Christiansen2024}\cite{Yohanan2012}\cite{Zheng2024psycho}.

Despite the growing use of soft robotics in performative and artistic contexts, most existing installations convey emotion primarily through visual or auditory signals, with a recent surge in tactile interaction. This haptic line, through controlled surface textures, deformation, or stiffness modulation, offers a direct and embodied channel of engagement, enabling audiences to feel the state of a robotic artwork rather than only observe it.

\begin{figure}[t] 
    \centering
    \includegraphics[width=\columnwidth]{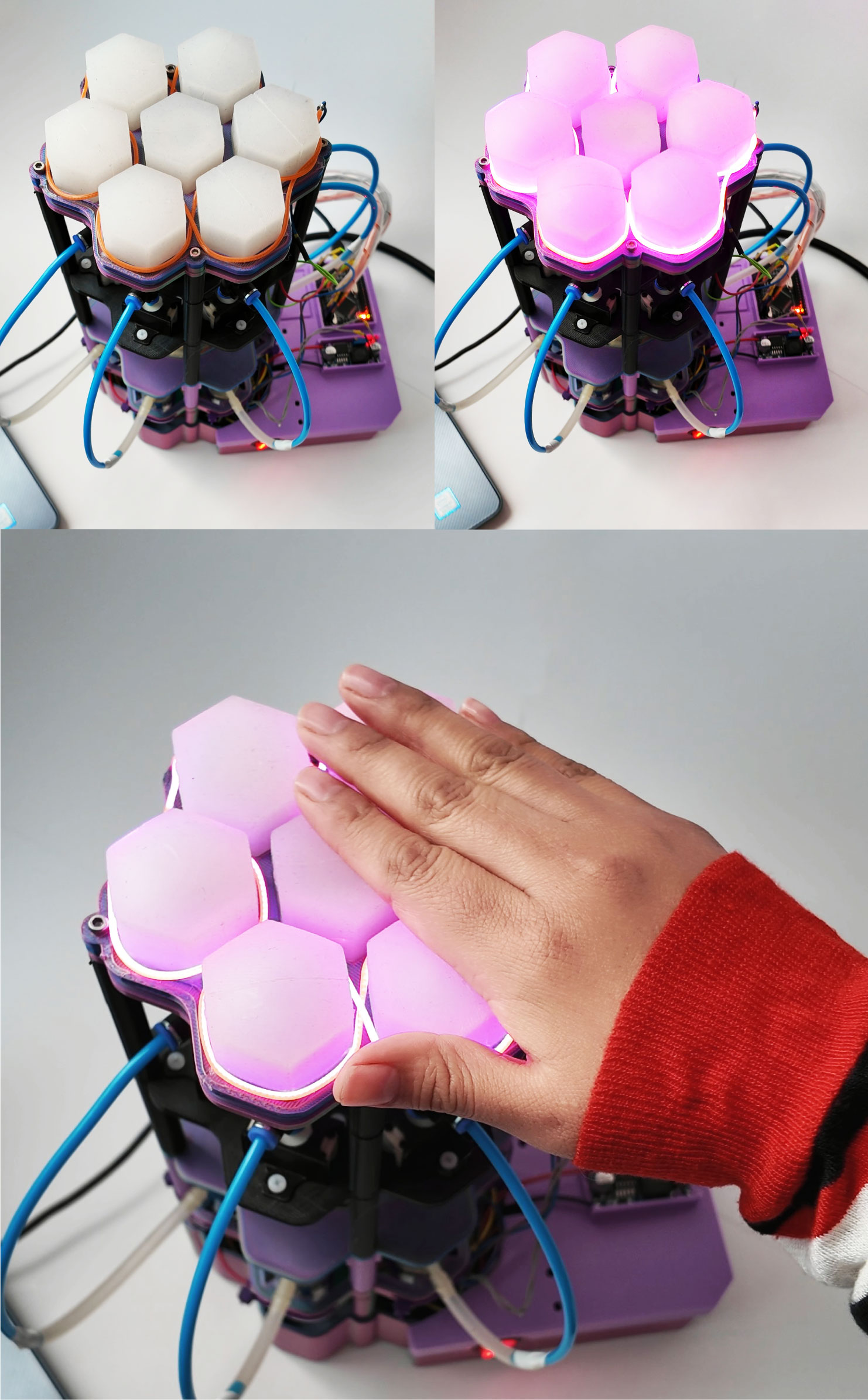} 
    \caption{Image of the interactive soft robotic installation. Top left: installation at rest. Top right: six MRing modules activated with maximum visual intensity. Bottom: user interacting with the active installation to experience the hardness of the modules.}
    \label{intro}
\end{figure}

\begin{figure*}[t] 
    \centering
    \includegraphics[width=\textwidth]{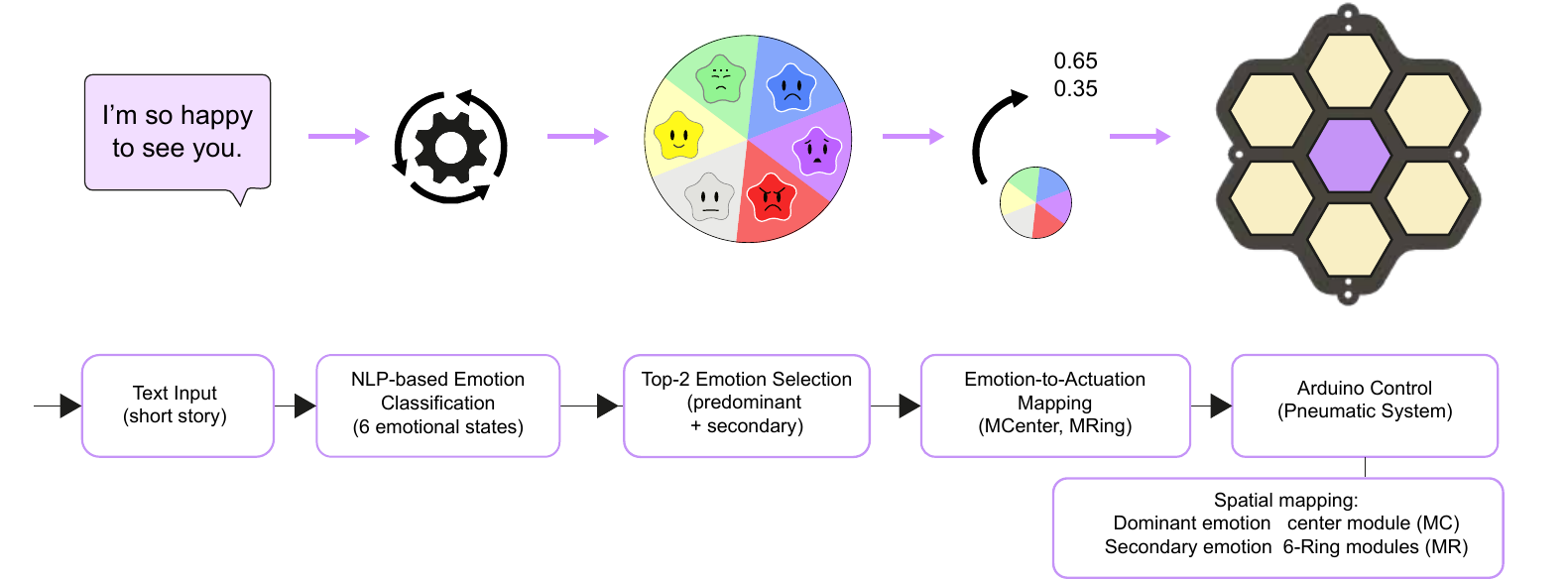} 
    \caption{General schematic of the proposed interactive soft robotics installation for the physical representation of emotions. The installation interprets short stories using a large language model (LLM) provided via the OpenAI API to infer the two most prominent emotions from a predefined set of six. The analysis outputs six emotional categories and identifies the top two predominant emotions, which are then used to generate specific pneumatic pressure levels for two groups of soft modules: MRing and MCenter.
}
    \label{diagrama_principal}
\end{figure*}

In this work, we introduce a variable-stiffness interactive soft robotics installation that maps semantic emotion analysis of narrative text to a distributed arrangement of soft pneumatic modules (See Figure \ref{intro}). The installation interprets short stories using a natural language model (OpenAI) to infer the two most prominent emotions from a predefined set of six. These emotions drive the inflation pattern of seven hexagonally arranged actuators, with the central module representing the primary emotion and the surrounding ring expressing the secondary one. Inflation levels are mechanically characterized to produce distinct stiffness states, enabling tactile expression of affect in real time and exploring stiffness modulation as an embodied channel for affective human–robot interaction.

We make the following contributions:
\begin{enumerate}
    \item We introduce a variable-stiffness soft robotic installation that, unlike prior works relying mainly on visual or auditory signals, explores stiffness modulation as an expressive and engaging pathway directly mapped from semantic emotion inference.

    \item We validate that a fully silicone with-membrane actuator enhances stiffness response while remaining low-cost and simple to fabricate, making it suitable for artistic contexts.
    \item We demonstrate how tactile interaction through variable stiffness enriches the aesthetic and experiential dimensions of soft robotics, advancing its role as both an artistic medium and a tool for public engagement.
\end{enumerate}

\section{Related Work}

The convergence of soft robotics, affective computing, and artistic expression has become increasingly evident in recent years. Spitale et al.~\cite{Spitale2025} review the rapid expansion of affective robotics for well-being, highlighting the growing integration of emotional and artistic dimensions in robotic systems. Christiansen et al.~\cite{Christiansen2024} introduce the concept of soft biomorphism, applying generative design to create complex, organic morphologies that merge functional and aesthetic qualities. Their approach emphasizes how material softness, structural curvature, and deformable geometries can evoke lifelike presence while maintaining mechanical performance—bridging engineering precision with artistic sensibility. In the artistic domain, Herath et al.~\cite{Herath2016} emphasize the potential of soft robotics for performative and interactive contexts, where compliance, deformability, and organic motion enable safe, tactile, and emotionally resonant audience engagement.

Touch has been widely recognized as a powerful channel for affective communication in human-robot interaction. Prior work has shown that humans naturally use touch to communicate emotional states to robots and expect emotionally meaningful tactile responses in return~\cite{Yohanan2012}. Beyond instrumental sensing, tactile qualities such as softness and surface properties strongly shape hedonic motivation to touch and perceived pleasantness, as captured by the concept of touch-ability~\cite{klatzky2012}. 

From a neurophysiological perspective, affective touch is supported by specialized sensory pathways that are distinct from purely discriminative touch and are closely linked to emotional processing and well-being~\cite{mcglone2014}. Together, these findings suggest that tactile interaction can function as an affective and experiential medium, rather than merely a perceptual input channel.

Recent research has explored robotic skin, surface texture, and shape change as expressive modalities for conveying affect. Hu and Hoffman~\cite{hu2019} \cite{hu2023} \cite{hu2018} introduced texture-changing pneumatic skins that use surface-level deformations such as goosebumps and spikes to express affective states in social robots, mapping continuous emotional dimensions (valence and arousal) to tactile and visual texture changes. Extending this line of work, Hu and Hoffman~\cite{hu2023} conceptualized robotic skin as an active expressive medium and proposed a taxonomy of texture units for affective interaction. Complementarily, Neto et al.~\cite{neto2024} showed that shape-changing robotic skins can elicit consistent associations with emotional categories such as happiness, anger, fear, calmness, and sadness through tactile perception, including in children with visual impairments. Together, these findings support the role of tactile and morphological variation as an expressive channel for emotion in HRI.

While prior work on expressive robotic skin focuses primarily on surface texture and micro-scale deformations driven by internal robot affective states, comparatively little attention has been given to volumetric stiffness modulation as an expressive modality. Moreover, most existing affective robotic systems aim to convey a robot’s internal emotional state or respond to human affect, rather than externalizing the emotional structure of an external narrative through tactile interaction.  

This leaves open how softness, resistance to touch, and compliance may support narrative emotion in embodied, multisensory, and artistic interactions. 

Art-based implementations of soft robots have also explored different sensory channels for affective expression. Shape-changing interfaces have been widely studied as a means of conveying internal states and engaging users through embodied interaction. 
For instance, Klausen et al.~\cite{klausen2022} present a soft robotic system that communicates emotional states through breathing-like motion, highlighting how rhythmic deformation can support affective interpretation without relying on explicit visual cues.

Similarly, Sprout~\cite{Koike2024}, a soft expressive robot, demonstrates a fiber-embedded soft actuator capable of extending, twisting, and expanding to convey internal states such as anger or curiosity. Silva-Plata et al.~\cite{SilvaPlata2025} integrate capacitive touch sensing with real-time mechanical simulation to achieve deformation-resilient multi-touch detection on soft sculptures, enabling interactive, tactile-rich experiences. While visually and conceptually compelling, these artistic systems primarily rely on visual and motion-based channels for affective communication, without grounding emotional expression in quantitative tactile mechanics or variable stiffness as an expressive signal.

Finally, prior work on embodied affect has shown that shape and motion alone can communicate emotional qualities even in abstract, non-anthropomorphic interfaces. Tan et al.~\cite{Tan2016} demonstrated that shape-changing interfaces can evoke affective interpretations based on principles of biological motion. Our work extends this perspective into the tactile domain by treating volumetric stiffness modulation and resistance to touch as embodied carriers of narrative emotion.

\section{Proposed Interactive Soft Robotics Installation}

The proposed installation consists of seven pneumatic soft modules with variable stiffness, designed to represent six emotional categories through changes in their mechanical response and color, making the experience perceptible both visually and haptically.

\begin{figure}[t] 
    \centering
    \includegraphics[width=\columnwidth]{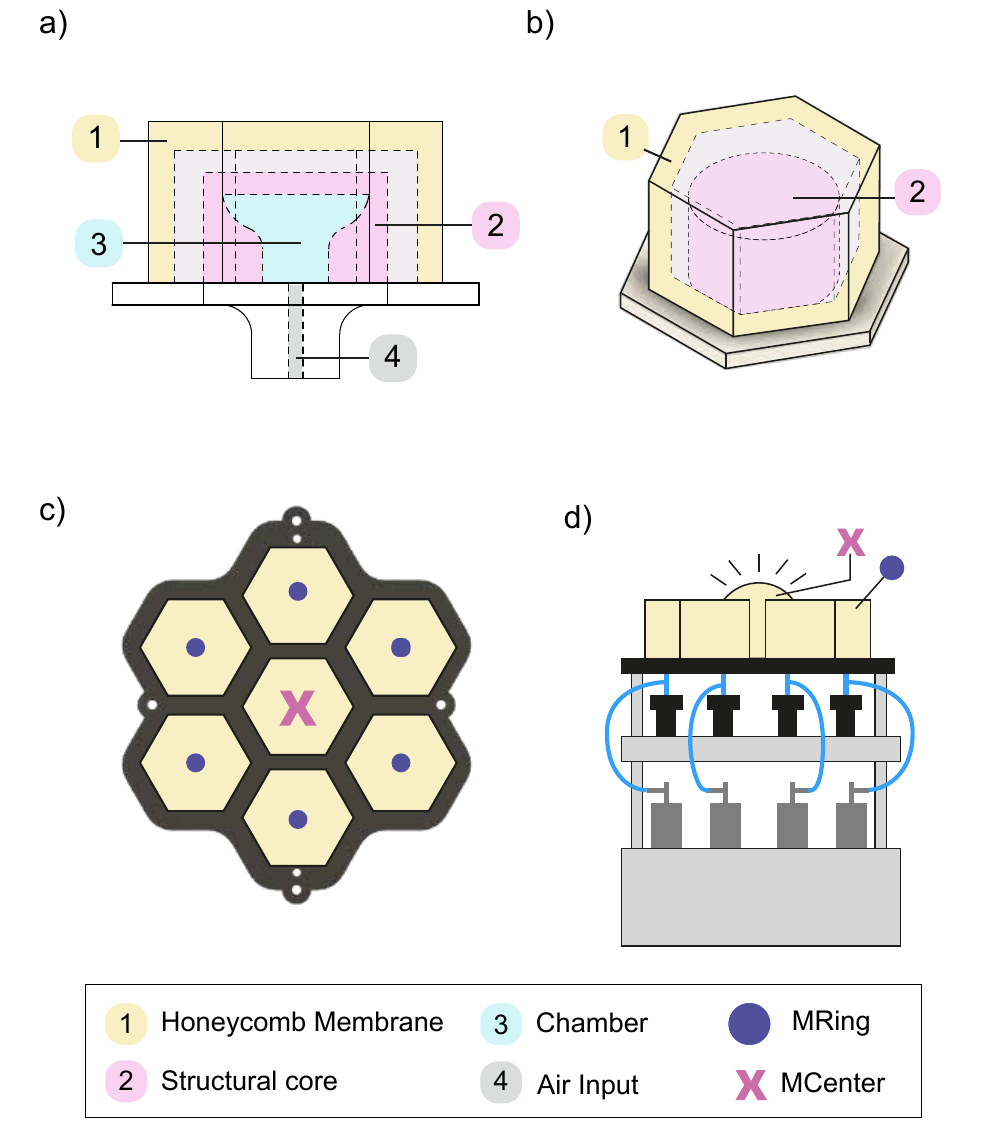} 
    \caption{a) Schematic view of the internal design of the soft module, consisting of a funnel-shaped inner chamber integrated within the main cylinder. This chamber is covered by a hexagonal-patterned membrane that is not directly attached to the cylinder wall, allowing controlled bulging during inflation. b) External view of the assembled soft module. c) Hexagonal arrangement of the installation: the six peripheral modules are referred to as MRing, while the central module is referred to as MCentral.  d) Complete interactive soft robotics installation view, including the microcompressors and solenoid valves responsible for inflating and deflating the modules, located on the top of the structure.}
    \label{modulo}
\end{figure}

The architecture comprises two main blocks:
\begin{enumerate}
    \item Soft module with membrane: A pneumatic cylinder covered by a hexagonal-pattern silicone membrane that is not directly bonded to the inner wall. This configuration allows additional control of bulging during inflation and, consequently, of the perceived stiffness. The membrane’s geometry was chosen to balance structural integrity and deformation freedom (See Figure \ref{modulo}.a and b.)
    \item Emotional pipeline: A text analysis model based on transformers (OpenAI API) processes the input text and identifies the two dominant emotions. Each emotion is mapped to a target internal pressure and an associated LEDs lighting intensity. The mapping was derived experimentally to ensure perceptible differences in both tactile and visual output.

    Each detected emotion is associated with a target pressure $P_t$ that produces a specific stiffness $k_t$, determined from prior mechanical characterization (next Section). This mapping allows emotions to be materialized in two sensory channels:
\begin{itemize}
    \item Visual: Through distinct intensity patterns.
    \item Haptic: Through tactile stiffness changes perceptible to the user.
\end{itemize}
Using two dominant emotions per input avoids ambiguity in emotional classification, reduces mapping complexity, and maintains a clear artistic narrative, particularly in interactive installations (See Figure \ref{diagrama_principal}).

\end{enumerate}

\section{Mechanical Characterization}
\subsection{Experimental Setup}

\begin{figure*}[t]
    \centering
    \includegraphics[width=\textwidth] {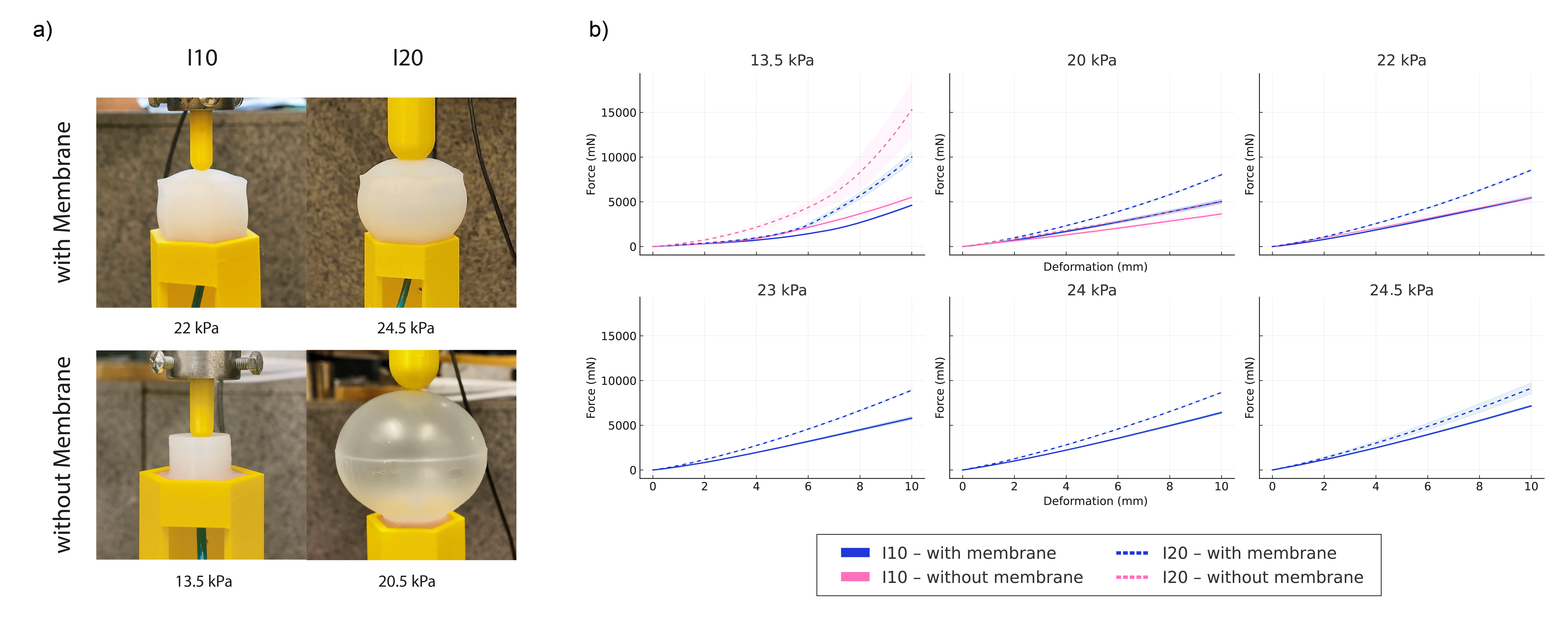}
    \caption{a) Experimental setup for indentation with $I_{10}$ and $I_{20}$ using a ZwickRoell Z100 and an Xforce~P \SI{1}{\kilo\newton} load cell, for both with membrane and without membrane configurations. b) Mean Force-Deformation curves at the evaluated inflation pressures; shaded regions denote the standard deviation (\(n{=}5\)).}
    \label{force_deformation}
\end{figure*}

We performed mechanical characterization of the soft inflatable modules using a ZwickRoell Z100 universal testing machine equipped with an Xforce~P load cell (nominal force: \SI{1}{\kilo\newton} ZwickRoell), enabling precise force acquisition throughout the experiments (Figure \ref{force_deformation}.a). We performed uniaxial indentation with two spherical indenters:
\[
I_{10}:~R=\SI{5}{\milli\meter}, \qquad
I_{20}:~R=\SI{10}{\milli\meter}.
\]
We tested two configurations of the soft inflatable modules to assess the effect of surface constraint on controllable stiffness: with membrane and without membrane. For each configuration and indenter, the module was quasi-statically inflated to target pressures
\[
p \in \{13.5,~20,~21,~22,~23,~24,~24.5\}\,\si{\kilo\pascal},
\]

Repeated \(n=5\) times per condition and indenter.

\subsection{Force-Deformation Curves and Stiffness}
The Force-deformation data \(F(\delta)\) were recorded for each pressure level (Figure \ref{force_deformation}.b). The initial stiffness \(k\) (in \si{\newton\per\milli\meter}) was computed from the linear region of the curve for small indentations (\(\delta \le \SI{1.0}{\milli\meter}\)). In practice, we obtained \(k\) as the least--squares slope,
\begin{equation}
k \;=\; \frac{\sum_{i=1}^{n} (\delta_i - \bar{\delta})(F_i - \bar{F})}{\sum_{i=1}^{n} (\delta_i - \bar{\delta})^2},
\label{eq:k_ols}
\end{equation}
and report, for each pressure \(p\) and condition, the mean stiffness \(\bar{k}(p)\) and the standard error of the mean (SEM).

Modules with membrane consistently tolerated higher pressures (up to \SI{24.5}{\kilo\pascal}) than the ones without membrane modules (limited to about \SI{20.5}{\kilo\pascal}), which showed larger volumetric expansion and mechanical hysteresis. The membrane constrains bulging and increases contact stiffness, yielding more robust and repeatable force–deformation responses (See Figure \ref{stiffness_young}).

\subsection{Apparent Young’s Modulus \texorpdfstring{$E^\ast$}{E*} (Hertz Contact)}
\begin{figure}[t]
    \centering
    \includegraphics[width=0.95\columnwidth]{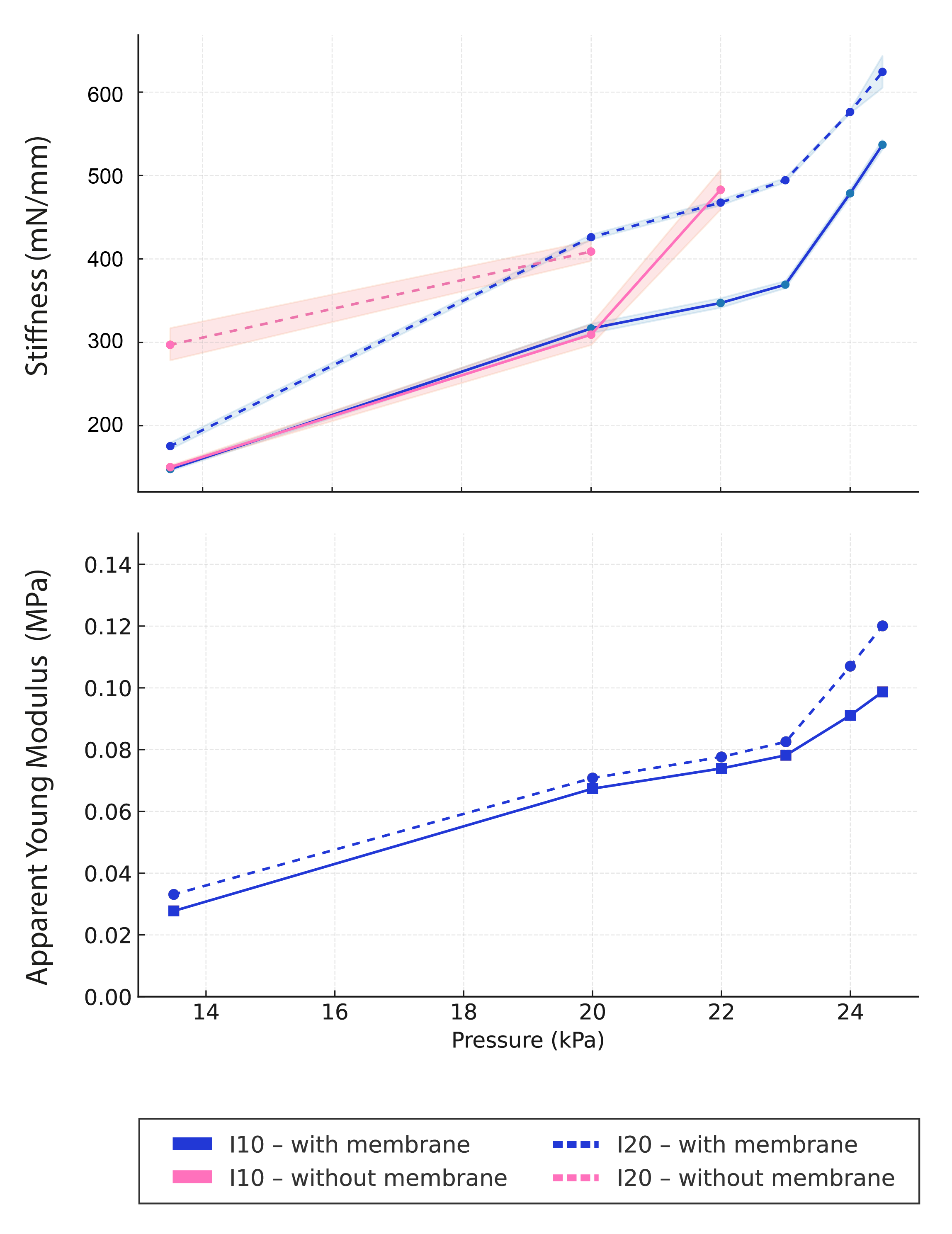}
    \caption{Top: Stiffness \(\bar{k}(p)\) (mean\(\pm\)SEM) vs.\ inflation pressure for both indenters and configurations. Bottom: Apparent Young’s modulus \(E^\ast(p)\) (mean\(\pm\)SEM) estimated via Hertz fitting. Modules with membrane exhibit systematically higher \(k\) and \(E^\ast\).}
    \label{stiffness_young}
\end{figure}

Assuming frictionless contact of a rigid sphere on a soft elastic half space, the Hertz model gives
\begin{equation}
F(\delta) \;=\; \frac{4}{3}\,E^\ast\,\frac{\sqrt{R}}{1-\nu^{2}}\,\delta^{3/2},
\label{eq:hertz}
\end{equation}
where \(F\) is force (\si{\newton}), \(\delta\) indentation depth (\si{\meter}), \(R\) indenter radius (\si{\meter}), and \(\nu\) Poisson’s ratio (silicone assumed \(\nu{=}0.49\)). For each condition, \(E^\ast\) (reported in \si{\mega\pascal}) was obtained by nonlinear regression of the experimental \(F(\delta)\) data to~\eqref{eq:hertz}. The resulting trend \(E^\ast(p)\) is shown together with \(\bar{k}(p)\) in Figure \ref{stiffness_young}, evidencing monotonic increases of both metrics with pressure.

\subsection{Design Innovation}

We propose a morphological control layer, a silicone membrane that enables dual-scale tuning of mechanical response (pressure and surface constraint). With this addition, we extend the operational range. Without membrane modules, the system expands volumetrically and exhibits viscoelastic hysteresis that we can exploit for expressive or dissipative interactions, whereas with membrane modules sustains higher pressures (up to \SI{24.5}{\kilo\pascal}) with improved rigidity control. Through this design, we establish a new trade-off space between volumetric compliance and surface-constrained stiffness. Beyond pure mechanics, we show how this trade-off informs the design of haptic interfaces for emotional communication, where both softness and rigidity contribute to the expressiveness of interactive soft robotics.


\section{Emotional Interactive Pipeline}

\begin{figure}[t] 
    \centering
    \includegraphics[width=\columnwidth]{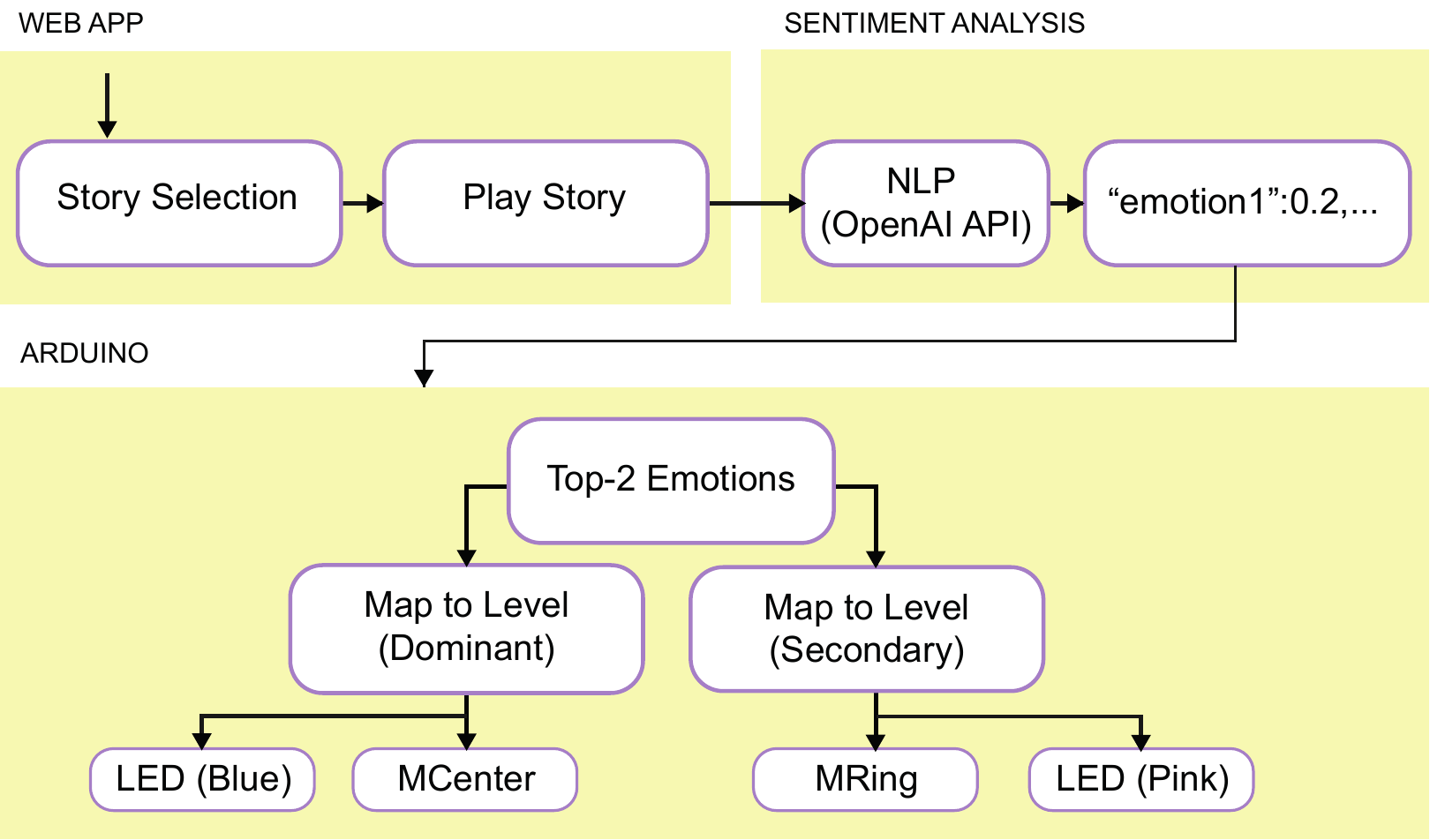} 
    \caption{Overview of the emotional interaction pipeline, illustrating the flow from paragraph reading and real-time sentiment analysis to the assignment of dominant and secondary emotions, fixed mapping of inflation levels, and parallel actuation of the MCenter and MRing module groups through pneumatic and visual signals.}

    \label{emotion_pipeline}
\end{figure}

\begin{figure*}[t] 
    \centering
    \includegraphics[width=\textwidth]{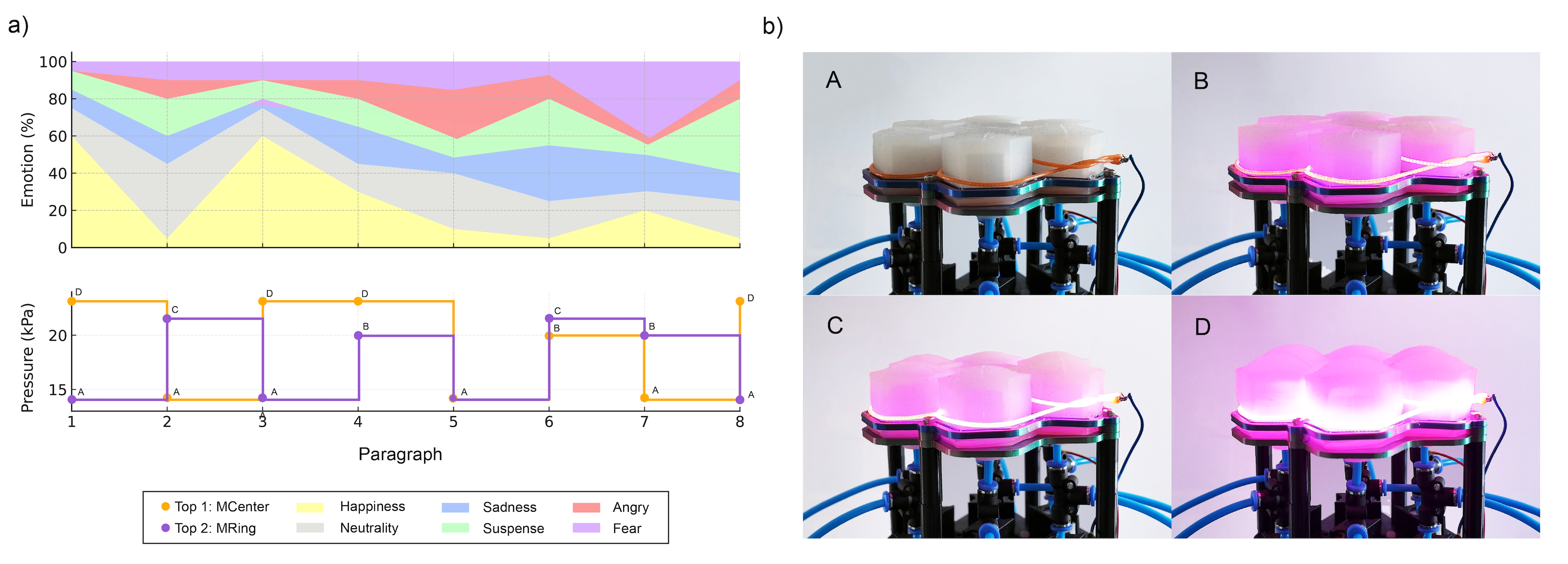} 
    \caption{Example of reading a story with eight paragraphs. a) Top: distribution of emotions. Bottom: inflation level assigned to MCenter and MRing according to the top-2 predominant emotions. b) Real images of the installation with each level in MRing}
    \label{emociones_ejemplo}
\end{figure*}
\begin{table}[h]
\centering
\caption{Inflation levels mapped to Young’s modulus $E^*$ and material analogies}
\begin{tabular}{lccc}
\hline
\textbf{Level} & \textbf{Pressure (kPa)} & \textbf{$E^*$ (MPa)} & \textbf{Material analogy} \\
\hline
A & 13.5 & 0.033 & Marshmallow~\cite{pestkaMarshmallow} \\
B & 20.0 & 0.070 & Gummy bear~\cite{williamsGummyBear}\\
C & 22.0 & 0.078 & Bread crumb~\cite{zghalBreadCrumb} \\
D & 23.5 & 0.095 & Gelatin hydrogel~\cite{czernerGelHydrogel} \\
\hline
\end{tabular}
\label{inflation_levels}
\end{table}

\begin{table}[]
\centering
\caption{Mapping of inflation levels for MRing and MCenter soft modules according to the detected emotion}
\begin{tabular}{lllll}
\hline
\multirow{2}{*}{\textbf{Emotion}} & \multicolumn{2}{c}{\textbf{Level Inflate}} & \multicolumn{1}{c}{\textbf{\begin{tabular}[c]{@{}c@{}}LED Blue\\ Intensity\end{tabular}}} & \multicolumn{1}{c}{\textbf{\begin{tabular}[c]{@{}c@{}}LED Pink\\ Intensity\end{tabular}}} \\
                                  & \textbf{MCenter}      & \textbf{MRing}     & \multicolumn{1}{c}{\textbf{MCenter}}                                                      & \multicolumn{1}{c}{\textbf{MRing}}                                                        \\ \hline
\textbf{Sadness}                  & B                     & B                  & Dim Light                                                                                 & Dim Light                                                                                 \\
\textbf{Anger}                    & D                     & A                  & Very Bright                                                                               & OFF                                                                                       \\
\textbf{Neutrality}               & A                     & A                  & OFF                                                                                       & OFF                                                                                       \\
\textbf{Happiness}                & D                     & D                  & Very Bright                                                                               & Very Bright                                                                               \\
\textbf{Suspense}                 & D                     & C                  & Very Bright                                                                               & Medium brightness                                                                         \\
\textbf{Fear}                     & A                     & B                  & OFF                                                                                       & Dim Light                         \label{level_inflate}                                                       
\end{tabular}
\end{table}

The proposed emotional interaction pipeline connects real-time text analysis with the mechanical and visual actuation of the soft modules. Each short story is divided into 6–14 small paragraphs. For our test, these stories were created by ChatGPT. We developed a local web interface to select the story and guide the reading process, display each paragraph in real time, and trigger the actuation pipeline. As soon as a paragraph is displayed, its text is sent to the OpenAI API for sentiment analysis.

The sentiment model detects six emotional categories, producing normalized scores whose sum equals 1. This string result is sent to the Arduino MEGA 2560, which determines the two dominant emotions. The primary emotion always controls the MCenter soft module, while the secondary emotion is assigned to the MRing group (as shown in Figure \ref{modulo}.c). This fixed mapping ensures consistent emotional-tactile correspondence across stories (see Figure \ref{emotion_pipeline}).

The six emotional categories used in this installation are sadness, fear, happiness, suspense, neutrality, and anger. Four of these (sadness, fear, happiness, and anger) correspond to basic emotions that are widely recognized across cultures~\cite{ekman1992argument}. Neutrality was included as a baseline affective state commonly found in affective computing and sentiment analysis datasets~\cite{mohammad2015} and suspense is introduced as an anticipatory emotional state characterized by a combination of anticipation and fear~\cite{lehne2015}.

We chose a discrete set of emotional categories rather than a continuous affect model such as the valence–arousal circumplex, as this choice better aligns with the goals of a tactile, public-facing artistic installation. Discrete emotions support stable and repeatable mappings to stiffness-based tactile feedback and reduce ambiguity in affective interpretation through touch. This categorical approach also offers a more accessible and intuitive interaction for non-expert users. Although continuous models enable finer emotional gradations, discreteness was favored here to emphasize narrative readability and tactile clarity in an exploratory setting.

We selected four inflation levels as a compromise between mechanical expressiveness and perceptual discriminability. Increasing the number of levels would result in stiffness variations that are too subtle to be reliably distinguished through touch, reducing the robustness of communicating through tactile feedback. Prior work in haptic perception shows that discretely spaced mechanical intervals improve perceptual reliability and help mitigate sensory fatigue \cite{Jones2013PsychophysicalHaptics}. In this context, we choose four levels to enable the installation to represent a continuum from a fully deflated (rest) state denominated (A), to maximum stiffness (D), maintaining perceptible differences both visually and through touch, an essential feature for artistic installations involving soft robotics and human–robot interaction. Table \ref{inflation_levels} summarizes the inflation levels along with their corresponding pressures (\si{\kilo\pascal}), effective Young’s modulus $E^*$ (\si{\mega\pascal}), and an analogy to a familiar material of comparable stiffness. Figure \ref{emociones_ejemplo}.b illustrates how the real interactive soft robotics installation actuates with each inflation level.

To reduce latency, the installation caches emotion results for previously processed paragraphs, decreasing the response time from 2–8 seconds to virtually instant feedback for repeated text. The actuation commands (pressure setpoints and LEDs intensity) are processed in the Arduino, which regulates pneumatic valves and lighting in parallel.

Figure \ref{emociones_ejemplo} illustrates the complete process, from a story with eight paragraphs, reading and emotion detection to the allocation of pressures for each module group, and the corresponding deformation response for each level (See Table \ref{level_inflate}). It provides an overview of the data flow and highlights the integration between software-based emotion analysis and hardware-based tactile and visual expression.

Supplementary video: Demonstrates the interactive workflow of the installation, including story selection, emotion inference, module inflation, and tactile exploration.

\section{User Experience Evaluation}

\noindent\textbf{Evaluation Scope and Framing:}
The main goal of this study is to examine whether users perceive the tactile and visual responses of the installation as emotionally meaningful, and whether they feel a qualitative “match” between the emotions suggested by the narrative and the tactile sensations they experience. It is not to precisely measure how accurately emotions are recognized, nor to build a detailed psychophysical model of tactile perception. Instead, it serves as an exploratory and experiential evaluation.

This approach follows prior work on affective touch and experiential human–robot interaction, which focuses on the hedonic (how it feels) and semantic (what it means) aspects of touch, rather than on fine perceptual discrimination or sensory thresholds.

To assess the perceptual and interactive qualities of the installation, a short user study was conducted with ten volunteers. Participants were invited to interact freely with the soft robotics installation, in which the actuation of the modules followed the predefined emotion–inflation mapping. Immediately after the experience, they completed an anonymous questionnaire evaluating five dimensions: clarity of interaction, accuracy of emotion–actuation mapping, engagement, multisensory impact, and overall satisfaction.

Each dimension was rated on a nine-point Likert scale ranging from 1(strongly disagree) to 5 (strongly agree), with increments of 0.5. This evaluation provided both qualitative impressions and quantitative scores, enabling an initial understanding of the installation’s experiential and affective impact on non-expert users.

This scale was chosen to allow participants to express nuanced subjective judgments, while keeping the evaluation simple and easy to interpret in this exploratory setting.

The results, summarized in Table \ref{tab:user}, highlight consistent positive feedback across dimensions, with particularly high ratings for engagement and multisensory integration.

\begin{table}[t]
\centering
\caption{User experience ratings for the soft robotics installation.}
\label{tab:user}
\begin{tabular}{lccc}
\toprule
Dimension & Mean & SD & $n$ \\
\midrule
Clarity of interaction        & 4.6 & 0.5 & 10 \\
Emotion--actuation mapping   & 3.5 & 0.6 & 10 \\
Engagement                   & 4.9 & 0.3 & 10 \\
Multisensory impact          & 4.4 & 0.5 & 10 \\
Overall satisfaction         & 4.8 & 0.5 & 10 \\
\bottomrule
\end{tabular}
\end{table}

\noindent \textbf{Ethical Considerations:}
This study is exempt from formal ethical review. It involved a voluntary, anonymous survey with ten adult participants, collecting subjective impressions of tactile interaction. No personal or sensitive data were collected, and no risk was involved. Informed consent was obtained from all participants prior to participation.This exemption statement appears in the \textit{User Experience Evaluation} section.

\section{Experimental Evaluation}
This interactive soft robotics installation uses a variable-stiffness soft module as its primary actuator, capable of expressing the emotion detected in each analyzed text paragraph.

To achieve stiffness modulation, we select a low-cost design, manufactured entirely from Ecoflex~00-30 silicone. The module consists of an internal cylinder with a chamber like a funnel that, when inflated, generates a dome on its upper surface. An external membrane, which is not in contact with the dome in its initial state, acts as an expansion limiter. This concentrates the pressure and produces a controlled change in stiffness. This mechanism is novel, simple, and cost-effective.

Mechanical characterization of the module was performed using indenters of \SI{10}{\milli\meter} and \SI{20}{\milli\meter} in width. For this study, we selected the \SI{10}{\milli\meter} width, as it approximates the average width of a finger during palpation, estimated from the ink print of an index finger on paper. The perceived hardness of the module ranged between that of a marshmallow and a hydrogel.

The system was evaluated in an interactive installation, where each paragraph of the story activated the predefined mapping of emotions to specific inflation levels of the soft modules. We defined two module groups: MCenter and MRing, assigning the dominant emotion to MCenter and the secondary emotion to MRing. This distribution enabled both visual and tactile differentiation of emotional responses.

Each inflation level was associated with a specific pressure range (in kPa) and an equivalent stiffness ($E^\ast$ in MPa), along with analogies to familiar materials to aid perceptual interpretation. These levels remained constant throughout the interaction, ensuring mechanical stability and repeatability in the user experience.

During testing, we observed perceptible differences in both visual deformation and tactile resistance across the levels, confirming that the selected four levels were sufficient to represent a continuous stiffness range without confusion. The combination of mechanical variation with LEDs lighting intensity reinforced emotional interpretation, particularly for strong emotions such as happiness.

Finally, installation latency remained low thanks to caching of previously processed analysis results, allowing smooth transitions between paragraphs without perceptible interruptions. The integration of text analysis, soft pneumatic actuator control, and visual feedback operated consistently, faithfully reproducing the emotional mapping defined in Table \ref{inflation_levels} and illustrated in Figure \ref{emociones_ejemplo}.

In addition to the mechanical and interactive evaluation, we made a user study with ten volunteers to assess subjective perception. After engaging with the installation, participants completed a short anonymous survey covering five dimensions: clarity of interaction, accuracy of the emotion-to-actuation mapping, engagement, multisensory impact, and overall satisfaction. Responses were collected using a nine-point Likert scale (from 1 strongly disagree to 5 strongly agree, in increments of 0.5).

The five questionnaire dimensions were designed as ad-hoc exploratory measures intended to capture experiential and affective qualities of tactile interaction, rather than to serve as psychometrically validated scales. They were not derived from existing questionnaires, but were defined to reflect key aspects of user experience relevant to this installation.

Their conceptual framing is consistent with prior work on affective touch and experiential human–robot interaction, which emphasizes hedonic, semantic, embodiment-related, and engagement-related dimensions of tactile experience, rather than fine-grained perceptual discrimination or sensory thresholds \cite{yohanan2011}\cite{klatzky2012}\cite{Cansev2021}\cite{Jorgensen2021}.

Given the artistic and exploratory nature of the installation, the goal of this evaluation was to obtain an initial qualitative and quantitative impression of user experience, rather than a clinically or psychometrically rigorous assessment.

The results (Table \ref{tab:user}) showed consistently positive evaluations, with particularly high scores in engagement and multisensory integration. This suggests that participants were able to perceive and interpret the emotional expressivity of the installation not only visually but also through touch.

\begin{figure}
    \centering
    \includegraphics[width=\columnwidth]{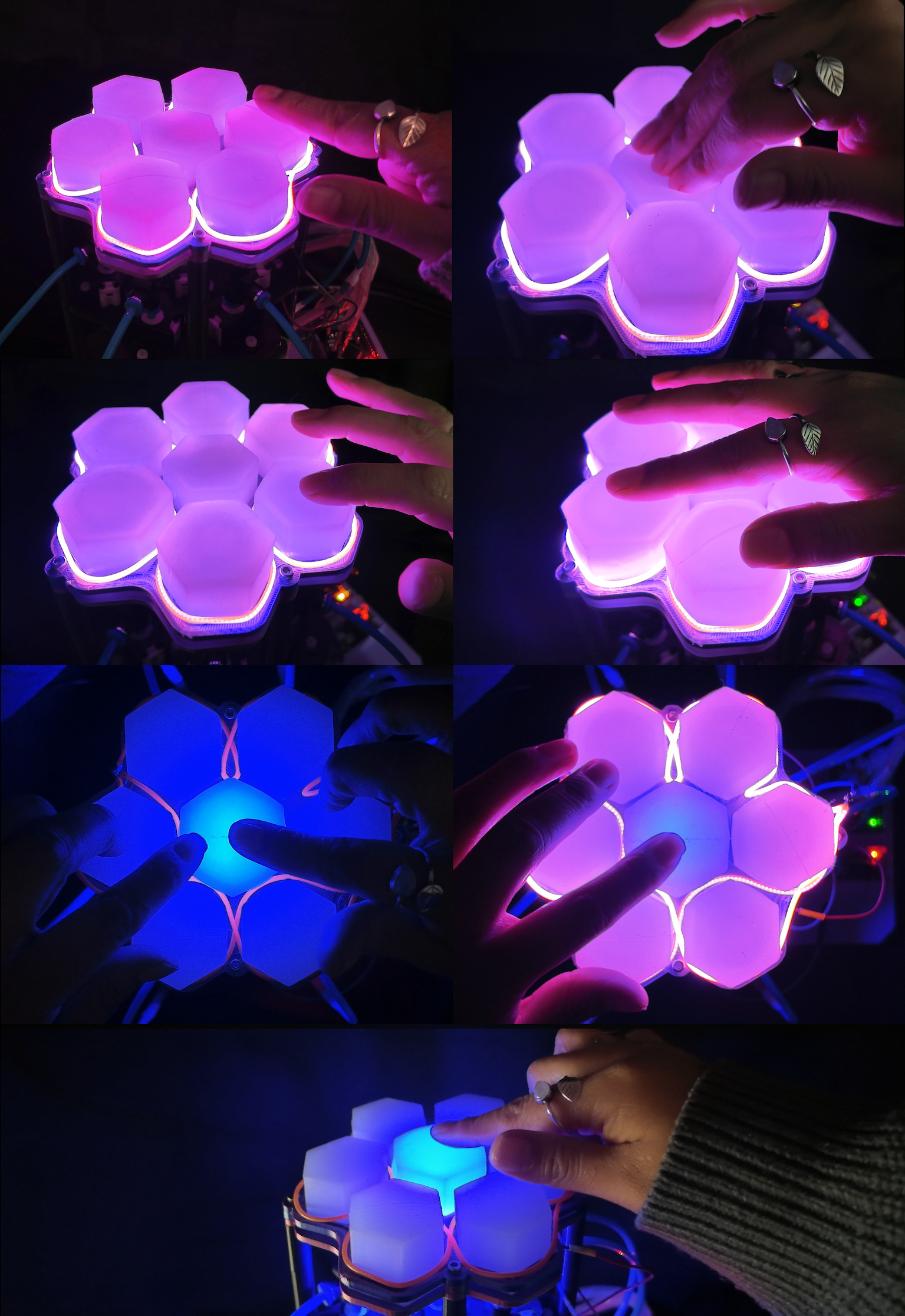} 
    \caption{Hands-on interaction with the soft robotic modules, which show the integration of touch, lights, and dynamic stiffness in a multisensory installation.}
    \label{user_experience}
\end{figure}

\section{Conclusions}

This work demonstrated the feasibility of coupling text to emotion recognition with soft robotic actuation in an interactive artistic installation. Our soft module enabled a novel and low-cost approach to variable stiffness control, with perceptible and reliable inflation levels that mapped emotions into both tactile and visual feedback (See Figure \ref{user_experience}).

The user evaluation reinforced these findings by showing that participants consistently perceived the intended emotional responses and valued the integration of mechanical and lighting effects. This indicates that beyond technical validation, the installation effectively engaged users in a multisensory manner, bridging engineering, art, and human–robot interaction.

By combining computational analysis, pneumatic actuation, and user-centered feedback, the study highlights the potential of soft robotics for creating emotionally expressive and perceptually rich interactive systems, enabling applications in art installations, therapeutic robotics, medical training, and more. 

Rather than proposing a new emotion model or aiming for precise emotion recognition, this work investigates the use of variable stiffness as an expressive component in embodied, narrative interaction. By combining text-driven emotion analysis with soft robotic actuation, the installation offers an exploratory perspective on how tactile properties such as softness and resistance to touch can contribute to emotionally meaningful experiences.

These findings align with prior work demonstrating that tactile and morphological variation can support emotional communication in human-robot interaction \cite{hu2023} \cite{neto2024} and that humans naturally interpret tactile interaction as emotionally meaningful \cite{yohanan2011}. In this sense, the installation can be understood as an embodied affective interface that extends visual and motion-based emotional expression into the tactile domain.

\section{Future Work}

Future work will focus on extending the adaptability and expressive range of the installation. This includes the development of dynamic control strategies for inflation levels and module grouping, as well as expanding text input to support freely authored narratives.

Additional actuation modalities, such as thermal or vibrotactile feedback, will be explored to strengthen multisensory engagement. The emotional categories may also be extended beyond the current six to support richer narrative interactions in both artistic and therapeutic contexts.

Finally, future studies will include larger-scale user evaluations using psychophysical and perceptual validation methods, such as confusion-matrix–based protocols and just-noticeable-difference (JND) measurements, to further assess the reliability and perceptual limits of stiffness-based emotional expression.

\bibliographystyle{IEEEtran}
\bibliography{references}

\end{document}